\documentclass[11pt]{article}

\usepackage[preprint]{acl}

\usepackage{times}
\usepackage{latexsym}

\usepackage[T1]{fontenc}

\usepackage[utf8]{inputenc}

\usepackage{microtype}

\usepackage{inconsolata}
\usepackage{booktabs}

\usepackage{graphicx}

\usepackage{amsmath,amssymb}
\usepackage{multirow}
\usepackage{xcolor}

\newcommand{\lab}[1]{\textit{#1}}

\newcommand{\BR}{BRIGHTER}
\newcommand{\GE}{GoEmotions}
\newcommand{\XED}{XED}
\newcommand{\ra}{$\rightarrow$}

\newcommand{\meansd}[2]{\shortstack{#1\\(#2)}}

\title{Fear Is Not Always Fear:\\Variation in Emotion Label Equivalence Across Datasets}
\author{Emily \"{O}hman \\
  Waseda University, Japan \\
  \texttt{ohman@waseda.jp} \\\And
  Anna Koufakou \\
  Florida Gulf Coast University, USA \\
  \texttt{akoufakou@fgcu.edu} \\}

\begin{document}
\maketitle


\begin{abstract}

Emotion classifiers are routinely applied beyond their training data, on the assumption that same-named labels such as \textit{fear} or \textit{surprise} mean the same thing across resources. We test this assumption for five labels shared by GoEmotions, XED, and BRIGHTER, comparing models on identical inputs and under cross-corpus transfer. On the same texts, one \textit{fear} classifier flags 71\% as fearful and another 8\%, yet the two largely agree on how the texts rank. Across 30 resource--label comparisons, training-label prevalence explains about half of this kind of scale mismatch, and a fitted intercept removes most of it. What recalibration cannot repair is larger: transferred models lose 55\% of native discrimination in aggregate, agreement in ranking between classifiers is only weakly related to that loss, and for \textit{surprise} the resources disagree about which texts express the emotion. The choice of training resource can therefore change conclusions: on two external tweet collections, BRIGHTER- and XED-trained classifiers give opposite answers about which shows more \textit{surprise}, even after rank normalization, with intervals excluding zero in both directions. A shared emotion label name does not establish an interchangeable measurement target.
\end{abstract}

\section{Introduction}

Emotion classifiers are increasingly used as measurement instruments: to track fear during a pandemic, compare joy across communities, or detect sadness in clinical text \citep[see e.g.][]{kleinberg-etal-2020-measuring,guo2024large,liu2024emotion,chu2025illusions}. A model fine-tuned on GoEmotions \citep{demszky-etal-2020-goemotions} or XED \citep{ohman-etal-2020-xed} is applied to new data, and its \textit{fear} scores are read as measurements of fear. This practice rests on an assumption that is rarely tested: that a label such as \textit{fear} refers to approximately the same target across datasets.

This assumption does not always hold, and the ways in which it fails have different implications for calibration fixes. We test it by training one model with a separate output head for each resource and comparing the heads on the same held-out texts. On the same 2,765 BRIGHTER test texts, the BRIGHTER \lab{fear} head classifies 70.8\% as fearful, averaged over three seeds, while the GoEmotions \lab{fear} head classifies only 7.6\%, a more than ninefold difference, using development-selected thresholds. Our prevalence analysis attributes much of this scale mismatch to base-rate differences, and the two heads largely agree on the \emph{rank ordering} of the texts by predicted \lab{fear} ($\rho=.897$): a study's absolute estimate of \lab{fear} can change dramatically with the training resource even when the models preserve nearly the same ranking of texts. For \lab{surprise}, by contrast, the XED and BRIGHTER heads disagree substantially about the ranking itself. The same headline observation, ``cross-resource disagreement'', therefore hides qualitatively different problems that require different remedies.

Emotion resources differ in theoretical grounding (fine-grained taxonomies versus Plutchik-style basic emotions; \citealt{plutchik1980psychoevolutionary}), inventory size, annotation guidelines, aggregation rules, and text domain: a label called \lab{fear} in a 28-category scheme competes with \lab{nervousness}, while in an eight-category scheme it does not. Such differences could move the category boundary, change how confidently a model predicts the label, or alter which texts are judged most representative of the emotion.

Existing work establishes that cross-corpus emotion classification is difficult \citep{bostan-klinger-2018-analysis} and that merging corpora into a unified inventory helps for some datasets but not others \citep{koufakou-etal-2024-towards}. What it does not establish is what form these cross-resource differences take. ``Poor transfer'' can arise from several distinct mismatches with very different practical consequences. A difference consistent with prior shift may in principle be corrected without target labels; a difference in probability scale can be corrected with a small number of them; a disagreement about which texts express the emotion cannot be fixed by any order-preserving recalibration.

This paper asks which of these mismatches actually occur between same-named emotion labels, focusing on \textit{anger}, \textit{fear}, \textit{joy}, \textit{sadness}, and \textit{surprise}. We make three contributions:

\begin{enumerate}
    \item \textbf{A framework for label equivalence} comprising four progressively weaker relationships: intercept, positive affine, and nonlinear monotonic equivalence, and ordering divergence.
    \item \textbf{A controlled same-input comparison.} A shared encoder with dataset-specific heads scores identical texts, yielding a 30-unit empirical map of cross-resource compatibility (five shared labels $\times$ six directed resource pairs).
    \item \textbf{An empirical characterization of portability and repairability} using independent cross-corpus transfer, progressive recalibration, a decomposition of probabilistic transfer error into miscalibration and discrimination loss, and corpus-level comparisons testing whether resource choice changes measurement conclusions.
\end{enumerate}

Same-named heads usually preserve substantial rank agreement while differing strongly in probability scale, and training-label prevalence explains much of that scale mismatch; transferred models nevertheless lose 55.3\% of native discrimination in aggregate, and \lab{surprise} shows ordering disagreement that no monotonic correction removes. These differences can change substantive conclusions: the direction of corpus-level comparisons depends on which resource the classifier was trained on, including for \lab{surprise} between two tweet collections external to all three training resources. We therefore argue that shared label names should be treated as hypotheses about equivalence, and that the relevant question is not whether two labels match but what form of equivalence they exhibit.\footnote{The code is available anonymously here: \url{https://anonymous.4open.science/r/label_harmony-7E4A/}}

\section{Related Work}
Emotion labels are shaped by the annotation scheme used to elicit them as well as by the underlying affective phenomenon. Annotations differ with the perspective annotators adopt \citep{buechel-hahn-2017-readers}, are entangled with intensity and annotator confidence \citep{troiano-etal-2021-emotion}, and depend on which aspect of an event is represented \citep{oberlander-etal-2020-experiencers}; corpora also differ in domain, scheme, and label inventory, and models trained on one do not transfer cleanly to another \citep{bostan-klinger-2018-analysis}. More broadly, annotation design is increasingly treated as part of the task definition rather than a neutral mechanism for recovering a pre-existing label: prescriptive and descriptive paradigms legitimately yield different label distributions \citep{rottger-etal-2022-two}, aggregation can obscure systematic variation in interpretation \citep{prabhakaran-etal-2021-releasing,davani-etal-2022-dealing}, and divergent emotion judgments are often legitimate rather than noise \citep{uma-etal-2021-semeval,plank-2022-problem,ohman-koufakou-2026-quality}. A dataset label is thus an operationalization of a construct, not evidence that the construct is measured consistently across resources \citep{bean2026measuring}. We make this distinction empirically testable for emotion labels by asking which forms of apparent equivalence (prevalence, probability scale, ordering) survive a change of annotation resource.

\citet{koufakou-etal-2024-towards} map heterogeneous emotion corpora into a common inventory and show that unified training has uneven effects across datasets. We take the complementary route: rather than collapsing resource differences, we retain each resource's own label function and ask which aspects of a nominally shared label are equivalent, and whether the differences are large enough to change corpus-level comparisons.

Relative judgments are more reliable than absolute rating scales in affect annotation \citep{kiritchenko-mohammad-2017-best,mohammad-bravo-marquez-2017-wassa,bagdon-etal-2024-expert}, so agreement in relative ordering need not imply agreement on an absolute scale; we extend this distinction from annotators to resource-level model behavior. Psychometrics draws a related distinction, under the heading of measurement invariance, between instruments that rank respondents alike and those that also share thresholds \citep{van2012checklist}; our analysis is operational, and we neither fit latent-variable models nor make formal invariance claims.

Platt scaling \citep{platt1999probabilistic} and isotonic regression \citep{zadrozny-elkan-2002} are standard post-hoc calibrators. Miscalibration under distribution shift is well documented \citep{guo2017calibration,ovadia2019trust}. EM-based prior adjustment \citep{saerens2002adjusting} corrects prevalence changes under the assumption that class-conditional distributions remain unchanged.We use these methods to study \emph{repairability}, i.e., how much transfer error progressively more flexible target-side correction removes, and the CORP score decomposition \citep{dimitriadis2021stable} to separate Brier-score differences into miscalibration and discrimination components.

\section{Levels of Label Equivalence}
\label{sec:framework}

Let $p_A(y\mid x)$ and $p_B(y\mid x)$ be the probabilities that two resources' models assign to a same-named binary label $y$ for text $x$, and let $\ell(\cdot)$ denote the logit. We distinguish four operational relationships between two such prediction functions. For the first three, the hierarchy is nested when the affine slope is positive and the nonlinear map is monotone increasing:

\begin{description}
    \item[L1: Intercept equivalence.] $\ell(p_B) = \ell(p_A) + b$. The two prediction functions differ only by a constant shift in log-odds. This relationship is consistent with a pure prior shift under the usual label-shift assumptions, but an intercept relationship alone does not establish that those assumptions hold.
    \item[L2: Positive affine equivalence.] $\ell(p_B) = a\,\ell(p_A) + b$ with $a>0$. The resources preserve text ordering but differ in probability scale or boundary sharpness.
    \item[L3: Nonlinear monotonic equivalence.] $p_B=g(p_A)$ for monotone-increasing $g$ (equivalently on logits). Ordering is preserved, but the probability scales are related nonlinearly.
    \item[L4: Ordering divergence.] No monotone-increasing $g$ adequately relates the two prediction functions: the resources disagree about \emph{which texts} receive relatively higher scores for the emotion.
\end{description}

These relationships represent progressively weaker forms of equivalence; we use them as reference relationships rather than assigning each empirical comparison to a discrete level. L1--L3 preserve the ordering of texts and differ only in how the two prediction scales are related, so rank agreement provides a complementary diagnostic; L4 permits the ordering itself to differ. Because an intercept relationship may arise from label-prevalence differences, we also compare the fitted L1 mapping with the log-odds shift implied by training-label prevalences. Target-side recalibration against gold labels (Section~\ref{sec:ladder-setup}) addresses a different question, repairability, and does not by itself assign a transfer unit to a level: residual error after isotonic recalibration is interpreted jointly with rank agreement, since residual error alone may also contain estimation noise.

\section{Experimental Setup}

\subsection{Data}
\label{sec:data}

\begin{table*}[t]
\centering
\small
\begin{tabular}{lrll}
\toprule
Resource & Items & Source data & Label scheme \\
\midrule
\GE{} \citep{demszky-etal-2020-goemotions} & 54,263 & Reddit comments & 27 emotions + neutral, multi-label \\
\XED{} (English) \citep{ohman-etal-2020-xed} & 27,202 & Movie subtitles & 8 Plutchik emotions + neutral \\
\BR{} (English) \citep{muhammad-etal-2025-brighter} & 5,643 & Social media & multi-label; intensity recoverable \\
SemEval-2018 E-c \citep{mohammad-etal-2018-semeval} & 10,981 & English tweets & 11 emotions; external evaluation only \\
\bottomrule
\end{tabular}
\caption{Training resources and external evaluation data. The cross-resource analyses focus on the five labels shared by all three: \lab{anger}, \lab{fear}, \lab{joy}, \lab{sadness}, \lab{surprise}.}
\label{tab:data}
\end{table*}

We use the three English resources in Table~\ref{tab:data}. The main analyses are restricted to the five labels that all three share (\lab{anger}, \lab{fear}, \lab{joy}, \lab{sadness}, \lab{surprise}). Multi-label items are retained as multi-label: each shared emotion is modeled as an independent binary target, so an item may be positive for more than one shared label. Only unambiguous aliases (e.g.\ \lab{happiness}\ra\lab{joy}) are normalized. Labels that a resource does not annotate are masked rather than treated as negatives. We preserve the released train/development/test partitions for \GE{} and \BR{}. Because the English \XED{} data did not provide equivalent partitions, we create a deterministic 80/10/10 split by SHA-256 hash of normalized text, ensuring that duplicate normalized texts remain in the same partition. The training/development/test partitions contain 43,410/5,426/5,427 examples for GoEmotions, 21,761/2,810/2,631 for English XED, and 2,763/115/2,765 for English BRIGHTER. 

\paragraph{External collections.}
For the external measurement comparison, we use the English emotion-classification (E-c) data from SemEval-2018 Task 1 \citep{mohammad-etal-2018-semeval}. The 2016 collection is the official training partition; the 2017 collection pools the official development and gold-labeled test partitions. Neither is used to fit or calibrate our models. After NFKC normalization, case folding, and whitespace normalization, we remove one text overlapping our models' training data and one within-collection duplicate, leaving 6,837 and 4,144 texts; no normalized texts are shared across the two collections. We compare these sampled collections rather than interpreting them as representative of year-to-year population change.

\subsection{Two model sources}

A standard transfer experiment confounds input domain, representation, prevalence, boundary, and semantics. We therefore use two complementary setups built from different models.

\paragraph{Compatibility (same-input comparison).}
A single encoder is trained jointly with one output head per resource, and every text is scored by all three heads, so the evaluation input and encoder representation are held fixed when same-named heads are compared. Remaining differences reflect heads learned under different supervision and training distributions; they should not be interpreted as isolating annotation semantics alone. We report Spearman correlation (ordering agreement) and mean absolute probability difference (MAE; scale agreement).

\paragraph{Portability (independent transfer).}

Separately, we train one native specialist per resource and apply each to the other two test sets, giving six directed transfers $\times$ five labels $=$ 30 transfer units. We report AUROC, average precision (AP; several target labels are sparse), F1, Brier score, and log loss. Excess error is measured relative to the native target specialist evaluated under the same procedure.

\noindent Because compatibility and portability come from different models, agreement between them is not an artifact of scoring a single model twice.

\paragraph{Implementation.}

All models use \texttt{FacebookAI/roberta-base} \citep{liu-etal-2019-roberta} as the pretrained encoder. Training uses AdamW with a learning rate of $2\times10^{-5}$, weight decay $0.01$, a linear learning-rate schedule with a warmup ratio of $0.06$, maximum sequence length 192, training batch size 16, and evaluation batch size 32. Models are trained for at most three epochs with early stopping (patience 1) and gradient clipping at a maximum norm of 1.0. Multi-label predictions use sigmoid outputs and binary cross-entropy over labels observed for the corresponding resource; labels not annotated by a resource are masked from the loss. The shared-encoder model uses natural sampling, with each training item seen once per epoch. All reported model results are averaged over three random seeds (13, 42, and 87).

\paragraph{Compute and software.}
The RoBERTa-base models contain approximately 125 million parameters each; the shared-encoder model with three classification heads contains approximately 126 million. Experiments used NVIDIA GH200 GPUs with 120\,GB of GPU memory. We estimate approximately 10 GPU-hours across training, exploratory runs, and evaluation, based on allocation records.

\subsection{Direct same-input mapping}
\label{sec:mapping-setup}

For each of the 30 directed resource--label comparisons, we fit L1--L3 mappings between same-named heads on identical target-domain development texts and evaluate them on held-out target-domain test texts, without using gold emotion labels. L1 fits an intercept-only logit mapping, L2 a positive affine logit mapping, and L3 a monotone isotonic mapping; all minimize squared error in probability space. The source-to-target training-prevalence log-odds shift serves as a parameter-free L1 baseline. We report held-out MAE and Spearman's $\rho$. To distinguish systematic disagreement from training variation, residual MAE is normalized by the larger of the source- and target-head same-head cross-seed MAEs, a conservative seed-stability reference.

\subsection{Recalibration ladder}
\label{sec:ladder-setup}

For each transfer unit, we fit target-side corrections on target development data and evaluate them on target test data: (i) intercept-only logistic recalibration; (ii) label-free prior-shift EM \citep{saerens2002adjusting}, applied both to raw outputs and to outputs first calibrated on source development data; (iii) Platt scaling; and (iv) isotonic regression. The native target model receives the same target-development calibration in each condition, so excess error compares transferred and native models under equivalent treatment. For thresholded evaluation, we select a separate label-specific threshold for each transferred and native model by maximizing F1 over all unique score cut-points on the target development set.

We additionally decompose raw Brier scores using the Consistent, Optimally binned, and Reproducible (CORP) decomposition \citep{dimitriadis2021stable} into miscalibration (MCB), discrimination (DSC), and uncertainty (UNC) components. Because transferred and native models are evaluated against the same target outcomes, the uncertainty term cancels, leaving excess miscalibration and loss of discrimination. We report discrimination loss both in absolute terms and relative to the native target model's discrimination, since absolute DSC is constrained by target prevalence. The PAV/isotonic fit used by CORP is estimated on the evaluation sample for diagnostic decomposition only; it is not a held-out recalibration procedure.

\section{Same Labels, Different Operational Behavior}
\label{sec:crosshead}
\begin{figure*}[t!]
    \centering
    \begin{minipage}[t]{0.49\textwidth}
        \centering
        \includegraphics[width=\linewidth]{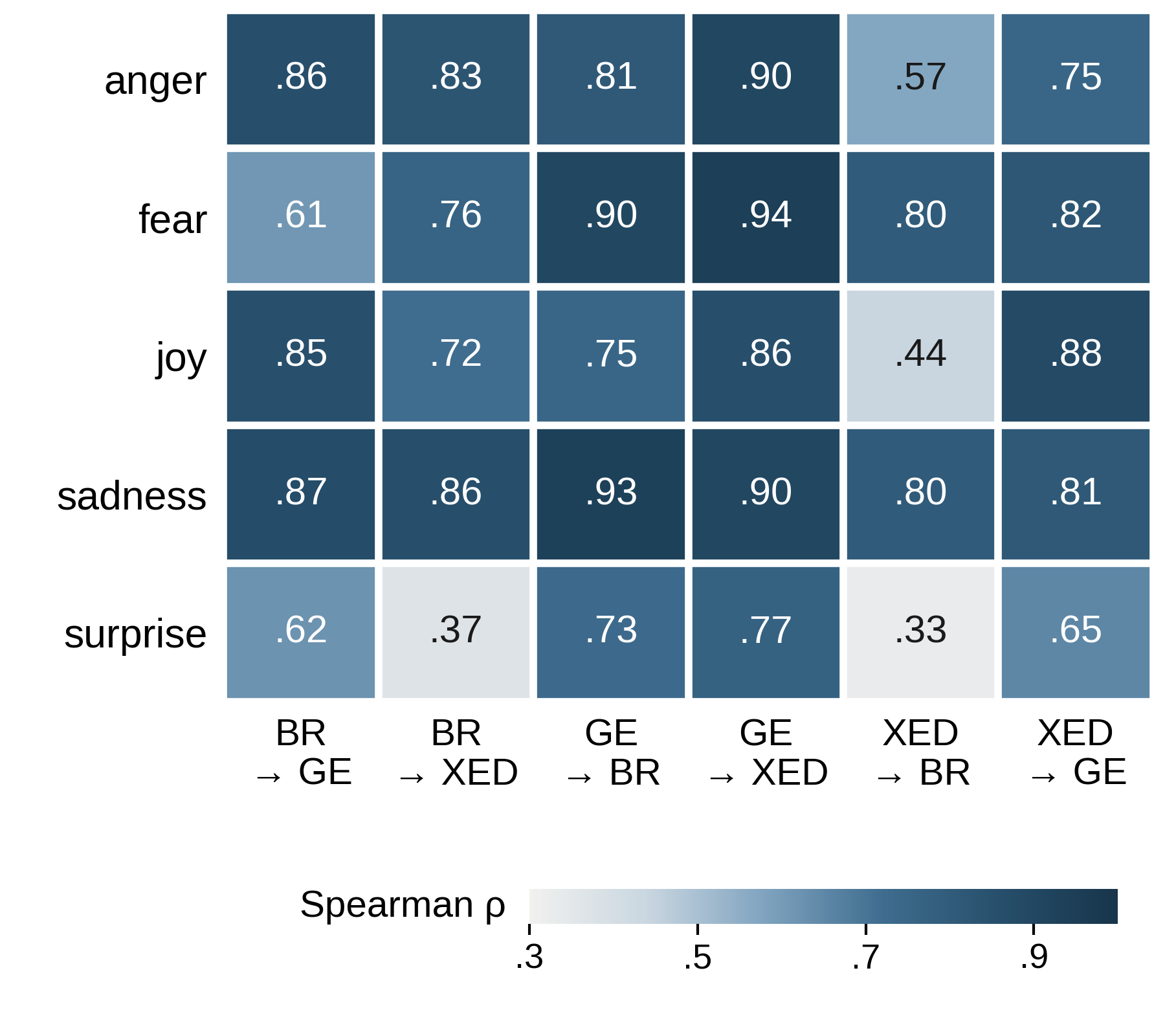}
        \small (a) Same-input rank agreement
    \end{minipage}
    \hfill
    \begin{minipage}[t]{0.49\textwidth}
        \centering
        \includegraphics[width=\linewidth]{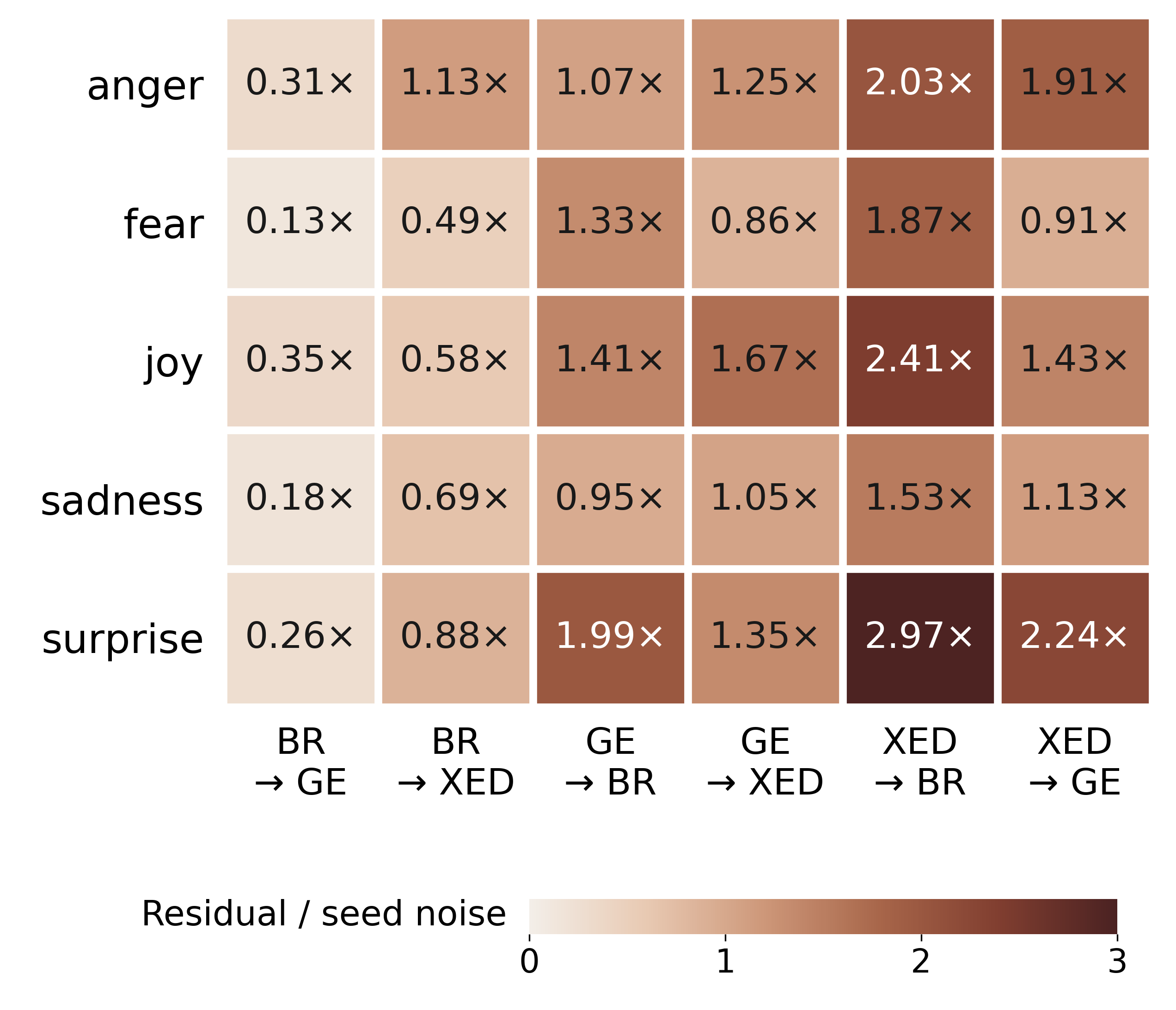}
        \small (b) Residual disagreement after monotonic mapping
    \end{minipage}

    \caption{Compatibility across the 30 directed resource--label comparisons. Panel (a) shows held-out same-input Spearman rank agreement. Panel (b) shows residual MAE after monotonic mapping, normalized by the larger of the source- and target-head same-head cross-seed MAEs. Values above 1 indicate residual disagreement larger than this conservative seed-stability reference.}
    \label{fig:same-input-examples}
\end{figure*}
\begin{table}[t]
\centering
\small
\begin{tabular}{lrr}
\toprule
Mapping & Mean MAE & Median MAE \\
\midrule
Raw & .193 & .184 \\
Prevalence intercept & .096 & .070 \\
L1: fitted intercept & .076 & .058 \\
L2: affine & .072 & .048 \\
L3: monotonic & .069 & .046 \\
\bottomrule
\end{tabular}
\caption{Held-out same-input mapping results across 30 directed resource--label comparisons. The prevalence-derived intercept uses only source and target training-label prevalences; L1--L3 are fitted on target-domain development predictions without gold emotion labels.}
\label{tab:crosshead}
\end{table}

Same-named heads (heads for the same-named label across resources) mostly agree on ordering and disagree on scale. Figure~\ref{fig:same-input-examples}a shows rank agreement for each directed resource--label comparison; across these 30 comparisons, mean agreement is $\rho=.756$ (median $.805$), and only three fall below $.50$. Table~\ref{tab:crosshead} summarizes probability-space mapping errors. A shift computed only from training-label prevalences cuts mean held-out MAE from $.193$ to $.096$ and improves 26 of 30 units;
a development-fitted intercept reduces it further to $.076$, improving on the prevalence shift in
20 of 30. Base rates therefore account for much, but not all, of the scale relationship. Affine and monotonic mappings add little on average. Figure~\ref{fig:same-input-examples}b shows the residual after monotonic mapping relative to the conservative seed-stability reference; these ratios average 1.21 (median 1.13).

\lab{Fear} illustrates a predominantly scale-based mismatch. On \BR{} texts, the \BR{} and \GE{} heads preserve high rank agreement ($\rho=.897$) despite a large probability-scale difference. The converse comparison on \GE{} texts has more moderate rank agreement ($\rho=.609$), but its raw MAE of $.377$ falls to $.012$ after monotonic mapping, within conservative same-head seed variation. Combining GoEmotions’ nervousness category with fear does not explain the mismatch with BRIGHTER: MAE stays at .517 under max pooling ($.517$) and reduces it only to $.507$ under noisy-OR, while slightly lowering rank agreement.

For \lab{surprise}, ordering itself diverges more
strongly (Figure~\ref{fig:same-input-examples}). The weakest comparison is \XED{}$\rightarrow$\BR{}, with $\rho=.329$; its monotonic residual remains 2.97 times the conservative same-head seed-stability reference. The reverse \BR{}$\rightarrow$\XED{} comparison also has weak rank agreement ($\rho=.368$), although its post-monotonic residual falls within same-head seed variation. This ordering divergence is consistent with L4-type incompatibility: an increasing monotonic mapping cannot reconcile different rankings. \lab{Joy} and \lab{anger} in \XED{}$\rightarrow$\BR{} likewise retain substantial residual disagreement after monotonic mapping (2.41 and 2.03 times the conservative reference, respectively). Same-named labels therefore do not share a single form of equivalence: probability-scale mismatch is widespread, while a smaller set of comparisons also exhibits substantial ordering divergence. A supplementary union-training comparison is reported in Appendix~\ref{app:union}.

\section{Cross-Corpus Portability}
\label{sec:transfer}

\subsection{Agreement vs.\ discrimination}

Table~\ref{tab:threshold} summarizes cross-corpus ranking and thresholded performance. With its own decision threshold, the \XED{} specialist reaches an F1 of .023 on \BR{}. With a threshold chosen on target development data, it reaches .562, against .702 for the native model, so much of the apparent zero-shot failure in this direction is an operating-point mismatch. Ranking ability transfers better than decisions: mean AUROC
across directions is .710--.855. Threshold selection is not sufficient everywhere: for \XED{}\ra\GE{}, mean F1 changes only from .171 to .169 and remains .399 below the native model.

\begin{table*}[t]
\centering
\small
\setlength{\tabcolsep}{3pt}
\begin{tabular}{lrrrrrrrr}
\toprule
Direction & AUROC & AUROC gap & AP & AP gap
& \shortstack{F1\\source thr.}
& \shortstack{F1\\target-dev thr.}
& Native F1 & F1 gap \\
\midrule
\BR\ra\GE &
\meansd{.855}{.0019} & \meansd{.095}{.0040} &
\meansd{.170}{.0092} & \meansd{.425}{.0130} &
\meansd{.115}{.0044} & \meansd{.243}{.0101} &
\meansd{.569}{.0105} & \meansd{.325}{.0202} \\

\BR\ra\XED &
\meansd{.786}{.0017} & \meansd{.126}{.0020} &
\meansd{.338}{.0028} & \meansd{.244}{.0050} &
\meansd{.319}{.0076} & \meansd{.391}{.0092} &
\meansd{.555}{.0008} & \meansd{.164}{.0098} \\

\GE\ra\BR &
\meansd{.773}{.0118} & \meansd{.115}{.0099} &
\meansd{.632}{.0104} & \meansd{.160}{.0099} &
\meansd{.253}{.0020} & \meansd{.582}{.0188} &
\meansd{.702}{.0141} & \meansd{.120}{.0259} \\

\GE\ra\XED &
\meansd{.753}{.0075} & \meansd{.159}{.0073} &
\meansd{.337}{.0035} & \meansd{.245}{.0088} &
\meansd{.210}{.0032} & \meansd{.363}{.0060} &
\meansd{.555}{.0008} & \meansd{.192}{.0067} \\

\XED\ra\BR &
\meansd{.753}{.0098} & \meansd{.135}{.0116} &
\meansd{.501}{.0126} & \meansd{.290}{.0126} &
\meansd{.023}{.0023} & \meansd{.562}{.0095} &
\meansd{.702}{.0141} & \meansd{.140}{.0108} \\

\XED\ra\GE &
\meansd{.710}{.0034} & \meansd{.240}{.0032} &
\meansd{.111}{.0100} & \meansd{.484}{.0175} &
\meansd{.171}{.0128} & \meansd{.169}{.0162} &
\meansd{.569}{.0105} & \meansd{.399}{.0240} \\
\bottomrule
\end{tabular}
\caption{Cross-corpus transfer averaged over five labels and three seeds. Parentheses give sample SD across seed-level five-label means. AP is average precision. Target-development and native F1 use target-development threshold selection; source-threshold F1 retains the source operating point. Gaps are native minus transferred performance.}
\label{tab:threshold}
\end{table*}

The weaker \XED{}\ra\BR{} results are not purely a property of difficult \BR{} labels. For example, \BR{} \lab{surprise} is well discriminated natively (AUROC $=.884$), yet AUROC falls to $.641$ from \GE{} and $.659$ from \XED{}. At the same time, \XED{}\ra\BR{} is broadly weaker for several labels, especially \lab{anger}. The \lab{surprise} result therefore combines a label-specific portability problem with a direction-level effect.

Heads can agree closely with each other and still fail to recover what target annotators marked. Of the 23 units with same-input rank agreement of at least $.70$, 15 still lose at least half of native discrimination, and across all 30 units rank agreement is only weakly related to discrimination loss ($\rho=-.275$; AUROC loss $\rho=-.215$).

\subsection{Probabilistic transfer error}
\label{sec:prob-transfer}

Probability-scale disagreement between compatibility heads closely tracks the excess Brier error of independently trained transferred models across the 30 units (Spearman $\rho=.973$). This correspondence is partly structural (Appendix~\ref{app:structural}), so we treat it as convergent evidence that the scale disagreement seen in the same-input comparison reappears as probabilistic transfer error, rather than as an independent predictive result.

The CORP decomposition shows that most excess probabilistic error is associated with miscalibration, but discrimination loss is also substantial. Averaged across the 30 transfer units, excess Brier score is $.112$: $.088$ is excess miscalibration and $.024$ is loss of discrimination. Relative to the native target model's discrimination, transferred models lose 55.3\% in aggregate (median unit: 63.0\%); 22 of 30 transfers lose at least half. The loss is especially pronounced for \lab{surprise}, for which 81.7\% of native discrimination is lost across transfers.

The native reference models are also not obviously limited by fixable training-label errors: LLM-assisted revision of flagged training labels yielded no consistent native improvement across three seeds (Appendix~\ref{app:llm-revision}).

\section{Recalibration Repair}
\label{sec:ladder}

\begin{table}[t]
\centering
\small
\begin{tabular}{lr}
\toprule
Correction & \makebox[0pt][r]{Mean excess Brier (SD)} \\
\midrule
None (raw)                       & .112 (.0020) \\
Prevalence-derived shift         & .055 (.0004) \\
Label-free prior-shift EM        & .122 (.0077) \\
Source-calibrated prior-shift EM & .125 (.0036) \\
Intercept-only                   & .031 (.0002) \\
Platt                            & .029 (.0003) \\
Isotonic                         & .023 (.0009) \\
\bottomrule
\end{tabular}
\caption{Mean excess Brier over 30 directed label-transfer
units; parentheses give the SD of these means across three
training seeds. The prevalence-derived shift uses training-label
prevalences. The native target model receives the corresponding
correction.}
\label{tab:ladder}
\end{table}

About half of probabilistic transfer error is a base-rate problem. A shift computed only from training-label prevalences halves mean excess Brier ($.112$ to $.055$) and improves 28 of 30 units (Table~\ref{tab:ladder}); a target-supervised intercept removes about 72\% of the raw excess ($.031$). Platt scaling provides a small additional reduction to $.029$, while isotonic regression reaches $.023$. Label-free EM correction instead worsens aggregate portability ($.122$ and $.125$). Its failure does not mean prevalence is unimportant: the direct prevalence shift helps substantially, though it does not close the transfer gap.

Directions nevertheless differ. For \BR\ra\GE{}, an intercept adjustment removes nearly all excess Brier error (raw $.172$, intercept $.008$, Platt $.008$, isotonic $.008$). For \XED\ra\BR{}, substantial residual mismatch remains (raw $.191$, intercept $.081$, Platt $.080$, isotonic $.047$). Thus, the effectiveness of an intercept varies considerably across resource pairs.

The number of target annotations needed is also target-dependent (Appendix~\ref{app:fewshot}): about 50 random annotations approach full-development performance for \XED{}, sparse \GE{} labels require more because small samples often contain no positives, and for \BR{} residual error changes little with sample size, indicating that its remaining gap is not a small-sample calibration problem.

\subsection{What remains after recalibration?}

\begin{table}[t]
\centering
\small
\setlength{\tabcolsep}{2pt}
\begin{tabular}{lrrrrrr}
\toprule
& & \multicolumn{4}{c}{Excess Brier} & \\
\cmidrule(lr){3-6}
Label & \shortstack{Same-\\input $\rho$}
& Raw & Intercept & Platt & Isotonic
& \shortstack{F1\\gap} \\
\midrule
\textit{fear} & .799 &
\meansd{.404}{.0012} & \meansd{.094}{.0009} &
\meansd{.098}{.0017} & \meansd{.039}{.0136} &
\meansd{.033}{.0617} \\

\textit{joy} & .445 &
\meansd{.143}{.0012} & \meansd{.080}{.0026} &
\meansd{.080}{.0027} & \meansd{.047}{.0103} &
\meansd{.122}{.0528} \\

\textit{surprise} & .329 &
\meansd{.166}{.0017} & \meansd{.097}{.0025} &
\meansd{.089}{.0015} & \meansd{.073}{.0019} &
\meansd{.206}{.0234} \\
\bottomrule
\end{tabular}
\caption{XED$\rightarrow$BRIGHTER: shared-head same-input rank agreement and excess Brier relative to the equally recalibrated native BRIGHTER model. The F1 gap uses exact target-development threshold selection. Parentheses give sample SD across three seeds for excess Brier and F1 gap.}
\label{tab:xed-brighter-residual}
\end{table}

XED\ra BRIGHTER carries the largest residual error, and its labels show that the residual is not of one kind (Table~\ref{tab:xed-brighter-residual}). \lab{fear} combines strong ordering agreement ($\rho=.799$) and a threshold-tuned F1 gap of only $.033$ with the largest raw excess Brier in the study ($.404$). An intercept adjustment removes most of this error ($.094$); Platt scaling does not improve it further ($.098$), while isotonic regression reduces the residual to $.039$.

\lab{surprise} shows the strongest ordering divergence in this direction. Same-input agreement is low ($\rho=.329$): ``My eye is black, my head hurts, but I'm happy, I won.'' falls near the 77th percentile of the \XED{} surprise ranking but the 5th of the \BR{} ranking, while ``After we moved, they stopped immediately.'' falls near the 89th percentile under \BR{} and the 16th under \XED{}. Even isotonic regression leaves substantial excess error ($.073$) and a threshold-tuned F1 gap of $.206$. The \XED{} head's ordering on \BR{} \lab{surprise} texts is substantially more stable across seeds (mean $\rho=.848$, range $.837$--$.864$) than its agreement with the \BR{} head, and \BR{} \lab{surprise} is not intrinsically difficult: its native AUROC is $.884$. Ranking each head separately on the same texts (seed-averaged predictions), 66 of 103 \BR{} test items in the top \BR{} and bottom \XED{} surprise quartiles are positive under the \BR{} gold labels (64.1\%); in the reverse group, only 1 of 28 is positive (3.6\%), compared with 28.9\% overall. The ordering disagreement therefore includes target-supported cases. 

Information lost when \BR{} annotations are aggregated does not independently explain the divergence: an initial association with transfer error disappears after matching items on their complete released five-label configuration (Appendix~\ref{app:compression}).

\section{Resource Choice Can Reverse Corpus Comparisons}
\label{sec:measurement}

To test whether these differences matter for downstream measurement, we compared mean emotion scores between each pair of held-out corpora under each of the three heads. The direction of the corpus comparison depends on the head in 8 of 15 corpus-pair--label comparisons. Five of these reversals disappear after converting each head's scores to pooled percentile ranks, indicating that they arise from probability-scale differences; three remain (Appendix~\ref{app:indomain}). Because these corpora are also the domains from which the models were learned, this analysis shows that measurements depend on the training resource, not how often such reversals would occur on arbitrary external corpora.

To examine whether reversals extend beyond the heads' training domains, we compared the external SemEval-2018 collections defined in Section~\ref{sec:data}. Table~\ref{tab:external} reports all five labels under all three heads. For \lab{surprise}, the mean pooled-rank difference (2016 minus 2017) is positive under the \BR{} head ($+.024$, 95\% item-bootstrap interval $[+.012,+.036]$) but negative under the \XED{} head ($-.015$, $[-.026,-.004]$); both signs hold in all three seeds. The \GE{} estimate is close to zero ($-.003$, $[-.014,+.009]$). Thus, a reversal occurs between collections external to all three training resources.

Across the five labels, raw mean differences have the same sign under all three heads. After rank normalization, sadness and surprise have opposing signs, but only surprise has opposite-direction intervals that both exclude zero; all heads rank the 2017 collection higher for joy, anger is close to zero, and fear is positive under all heads but clearly separated from zero only under \BR{}. Gold surprise prevalence differs by only $+.003$ ($[-.005,+.012]$), so the annotations do not establish which head gives the correct direction. The pointwise intervals quantify item-sampling uncertainty conditional on the fitted, seed-averaged predictions; they do not incorporate training variability or adjustment for multiple comparisons (Appendix~\ref{app:external}).

\begin{table}[t]
\centering
\small
\setlength{\tabcolsep}{4pt}
\begin{tabular}{lrrr}
\toprule
Label & \BR{} & \GE{} & \XED{} \\
\midrule
\textit{anger} & +0.011 & +0.012 & +0.001 \\
\textit{fear} & +0.038 & +0.009 & +0.006 \\
\textit{joy} & -0.036 & -0.046 & -0.050 \\
\textit{sadness} & +0.005 & -0.000 & -0.008 \\
\textit{surprise} & +0.024 & -0.003 & -0.015 \\
\bottomrule
\end{tabular}
\caption{External collection differences in mean pooled percentile rank (SemEval-2018 E-c; 2016 minus 2017), using seed-averaged predictions. All five shared labels and all three heads are shown. Full intervals and seed ranges appear in Appendix~\ref{app:external}.}
\label{tab:external}
\end{table}

\section{Discussion and Conclusion}

Prevalence and calibration account for most of how same-named emotion labels differ in level, but not for how they differ in ranking, and ranking differences are what survive the corrections a careful user would apply: they appear as lost discrimination after transfer and as a reversed comparison after rank normalization. A shared label name should therefore be treated as a hypothesis of operational equivalence, not evidence of it.

\paragraph{Practical recommendations.}
(1) Check whether training-prevalence differences explain an apparent scale mismatch before attributing it to deeper construct differences. (2) When target labels are available, begin with intercept-only recalibration; Platt scaling adds only modest average improvement, and required sample sizes are target-dependent, particularly for sparse labels. (3) Check both rank agreement on shared inputs and discrimination against target annotations: low rank agreement warns of ordering differences that monotonic recalibration cannot repair, while high agreement alone does not establish target-label validity. (4) Test whether substantive conclusions are robust to the choice of training resource.

\paragraph{Future work.} A formal measurement-invariance analysis with latent-variable models is a natural extension. \BR{} also annotates 28 languages under a single scheme, so the same-input design can test whether label equivalence holds across languages when guidelines and label inventory are held fixed.

\section*{Limitations}

We study three English resources, five labels, and one encoder family (RoBERTa-base); the pattern of relationships may differ for other languages, resources, or architectures. Our unit of analysis is small (30 directed label-transfer units drawn from three undirected dataset pairs), so correlations over these units are descriptive, and our robustness evidence rests mainly on per-seed replication and leave-one-out analyses. The same-input comparison holds evaluation texts and the shared encoder fixed, but the resource-specific heads are learned from different supervision and training distributions; it therefore shows that the heads differ, not whether the cause is annotation guidelines, label semantics, or domain. The L1--L4 framework characterizes relationships between prediction functions, while target-side calibration probes repairability and does not itself assign a transfer unit to a level. We analyze the five shared labels independently rather than modeling their multilabel structure jointly. The BRIGHTER compression analysis (see Appendix~\ref{app:compression}) is observational; its initial association with transfer error disappears after matching on the released five-label configuration.

\section*{Ethical Considerations}

This study is a secondary analysis of existing text annotations; we recruited no participants and collected no new personal data. We did not conduct an additional systematic audit for personal identifiers or offensive content, and do not claim that the source texts are anonymized or screened. Our planned release contains analysis code and aggregate results; source texts and item-level predictions are excluded, and users are directed to the original data providers.

Emotion predictions describe model responses to text under a particular annotation scheme. They should not be interpreted as individual psychological assessments or used for consequential decisions about individuals without separate validation. Dataset and model licenses are documented in  Appendix~\ref{app:resources}.

\bibliography{custom}

\section*{Acknowledgments}

This work was supported by JSPS KAKENHI Grant Number 24K21058.

Computational resources were provided by CSC -- IT Center for Science, Finland, through the Roihu computing environment.
\appendix




\section{Shared-Label Union Training}
\label{app:union}

\begin{table*}[t]
\centering
\small
\begin{tabular}{lrrr}
\toprule
Training condition & \BR & \GE & \XED \\
\midrule
Native specialist                       & .713 & .468 & .598 \\
Naive union (shared label heads)        & .672 & .418 & .591 \\
Shared encoder + dataset-specific heads & .715 & .432 & .595 \\
\bottomrule
\end{tabular}
\caption{Macro-F1 over each dataset's native label inventory, averaged over three seeds; each dataset is evaluated with its own development-set thresholds.}
\label{tab:union}
\end{table*}

If same-named labels are not fully operationally equivalent, forcing them onto a single output may reduce native performance. Table~\ref{tab:union} shows this pattern for \BR{} ($-.041$) and \GE{} ($-.050$), while the difference for \XED{} is negligible ($-.007$). Giving each resource its own head on a shared encoder recovers the \BR{} result and part of the \GE{} loss. Because union training also changes the data mixture and loss balance, these results do not isolate label incompatibility, but they are consistent with the direct same-input and transfer analyses above.

\section{Resource versions and terms of use}
\label{app:resources}

We use the filtered, agreement-based GoEmotions release with its official partitions, the English XED data with the text-hash partitions described in Section~\ref{sec:data}, and the English BRIGHTER emotion-category data. GoEmotions, XED, and the BRIGHTER emotion-category release are distributed under Creative Commons Attribution 4.0
(CC BY 4.0).\footnote{\url{https://github.com/google-research/google-research};
\url{https://github.com/Helsinki-NLP/XED};
\url{https://huggingface.co/datasets/brighter-dataset/BRIGHTER-emotion-categories}}

The external evaluation uses the English emotion-classification (E-c) files from SemEval-2018 Task~1: the official training file forms the 2016 collection, and the development and gold test files form the 2017 collection.
The NRC terms permit research use with attribution, require contacting the provider for commercial use, and prohibit redistribution of the source data.\footnote{\url{https://saifmohammad.com/WebPages/SentimentEmotionLabeledData.html}}

The pretrained checkpoint \texttt{FacebookAI/roberta-base} is distributed under the MIT license.\footnote{
\url{https://huggingface.co/FacebookAI/roberta-base}} The supplementary resource manifest records the
downloaded revisions, source filenames, and SHA-256 checksums.

The reproduction environment, checked using the \texttt{python-pytorch/2.13} module, comprised Python 3.12.14, PyTorch 2.13.0+cu130, Transformers 5.16.1, NumPy 2.5.2, pandas 2.3.3, SciPy 1.18.1, scikit-learn 1.9.0, datasets 5.0.1, and tokenizers 0.23.1.

\section{Structural Relationship Between Scale Disagreement and Excess Brier Error}
\label{app:structural}

Section~\ref{sec:prob-transfer} reports that probability-scale disagreement between compatibility heads closely tracks the excess Brier error of independently trained transferred models across the 30 units (Spearman $\rho=.973$; per seed $.964$--$.967$). This correspondence is partly structural. For source and target probabilities $p_s, p_t$,
\begin{equation}
(p_s - y)^2 - (p_t - y)^2 = (p_s - p_t)^2 + 2(p_s - p_t)(p_t - y).
\end{equation}
If $p_t(x)$ equals the target conditional probability $\mathbb{E}[Y\mid X=x]$, the cross-term has zero conditional expectation; ordinary marginal calibration alone is not sufficient for this cancellation. We therefore treat the observed correlation as convergent evidence that the probability-scale disagreement seen in the controlled same-input comparison reappears as probabilistic transfer error, rather than as an independent predictive result.

The relationship remains strong when leaving out any label ($\rho=.964$--$.981$) or dataset pair ($.940$--$.962$), for log loss ($\rho=.902$), and after controlling for training-prevalence difference (partial $\rho=.941$).

\section{Few-Shot Recalibration}
\label{app:fewshot}

\begin{table}[t]
\centering
\small
\setlength{\tabcolsep}{4pt}
\begin{tabular}{lrrrr}
\toprule
Target & $n=25$ & $n=50$ & $n=100$ & Full \\
\midrule
\BR{}  & .063/.057 & .063/.059 & .062/.060 & .062/.060 \\
\GE{}  & .059/.057 & .039/.034 & .019/.016 & .011/.008 \\
\XED{} & .025/.021 & .022/.019 & .021/.019 & .021/.019 \\
\bottomrule
\end{tabular}
\caption{Mean excess Brier under random few-shot recalibration. Entries are intercept/Platt.}
\label{tab:fewshot}
\end{table}

We repeat both intercept-only and Platt recalibration using target-development samples of 25, 50, and 100 items, together with the full development set, capped at the available development-set size. Random samples containing only one class fall back to the uncalibrated model; we additionally use a label-stratified diagnostic condition that guarantees both classes. Because sample requirements differ substantially by target dataset, results are reported by target rather than only as an aggregate.

Few-shot requirements differ sharply by target (Table~\ref{tab:fewshot}). For \XED{}, about 50 random target annotations approach full-development performance. Sparse \GE{} labels require larger random samples because small samples often contain no positives. When both classes are guaranteed, however, 25 \GE{} examples already approach full-development performance, showing that class coverage drives much of the sample requirement. For \BR{}, residual excess error changes little as the calibration sample grows, indicating that its remaining transfer gap is not primarily a small-sample calibration problem. Intercept-only recalibration follows the same qualitative pattern as Platt scaling.

\section{BRIGHTER Annotation-Compression Robustness Analysis}
\label{app:compression}

We examined whether information lost when individual \BR{} annotations are projected into the released label space could explain the comparatively large transfer discrepancies into \BR{}. Individual annotations were recovered for 5,623 of 5,643 English items (mean 8.5 raters per item). Among the recovered items, 49.0\% show at least one diametric affect pattern in the raw annotation space, compared with 38.2\% in the released representation. Mean pairwise label-set Jaccard agreement increases from 0.572 to 0.594 after projection. In total, 608 items (10.8\%) contain diametric structure in the raw annotations that is absent from the released representation; we refer to these as \emph{raw-only-diametric} items.

\paragraph{Unadjusted association.}
On the \BR{} test set, 296 of 2,765 items are raw-only-diametric. Table~\ref{tab:compression-unadjusted} compares their item-level excess transfer error with the remaining test items. Before recalibration, raw-only items show greater excess error for both \GE{}$\rightarrow$\BR{} and \XED{}$\rightarrow$\BR{}. The difference is no longer evident after Platt or isotonic calibration.

\begin{table*}[t]
\centering
\small
\begin{tabular}{lccc}
\toprule
Source & Raw & Platt & Isotonic \\
\midrule
\GE{}$\rightarrow$\BR{}
& $+0.0346$ [+0.0138, +0.0547]
& $+0.0039$ [-0.0067, +0.0144]
& $+0.0001$ [-0.0116, +0.0112] \\
\XED{}$\rightarrow$\BR{}
& $+0.0453$ [+0.0225, +0.0683]
& $-0.0034$ [-0.0152, +0.0078]
& $-0.0037$ [-0.0155, +0.0080] \\
\bottomrule
\end{tabular}
\caption{Difference in item-level excess squared error between raw-only-diametric and other \BR{} test items (raw-only minus other), averaged over the five shared labels. Brackets give 95\% bootstrap confidence intervals. Positive values indicate greater transfer error on the raw-only subset.}
\label{tab:compression-unadjusted}
\end{table*}

\paragraph{Controlling for released label structure.}
Raw-only-diametric items are not distributed randomly across the released label space, so the unadjusted association may reflect the final label configurations in which these items occur. We therefore performed two robustness analyses. First, we estimated the raw-only effect while controlling for the presence of the focal target label, the number of other positive released labels, emotion-label fixed effects, and, in pooled models, source fixed effects. Standard errors were clustered by \BR{} item. Second, we used
only within-stratum comparisons between items with exactly the same complete released five-label pattern. This exact-pattern analysis observed 28 released patterns, of which 15 contained both raw-only and comparison items. It retained 2,087 test items, including all 296 raw-only-diametric cases and 1,791 comparison items.

\begin{table*}[t]
\centering
\scriptsize
\setlength{\tabcolsep}{4pt}
\begin{tabular}{llccc}
\toprule
Control & Method & \GE{}$\rightarrow$\BR{} & \XED{}$\rightarrow$\BR{} & Pooled \\
\midrule
Label density / target presence & Raw
& $-0.0022$ [-0.0136, +0.0091] & $+0.0040$ [-0.0076, +0.0156] & $+0.0009$ [-0.0098, +0.0115] \\
& Platt
& $-0.0021$ [-0.0122, +0.0080] & $-0.0112$ [-0.0221, -0.0003] & $-0.0066$ [-0.0161, +0.0028] \\
& Isotonic
& $-0.0051$ [-0.0161, +0.0060] & $-0.0087$ [-0.0202, +0.0028] & $-0.0069$ [-0.0169, +0.0032] \\
\addlinespace
Exact released five-label pattern & Raw
& $+0.0028$ [-0.0079, +0.0135] & $-0.0004$ [-0.0112, +0.0105] & $+0.0012$ [-0.0087, +0.0111] \\
& Platt
& $+0.0049$ [-0.0051, +0.0148] & $+0.0004$ [-0.0104, +0.0112] & $+0.0026$ [-0.0067, +0.0119] \\
& Isotonic
& $+0.0033$ [-0.0072, +0.0138] & $-0.0046$ [-0.0162, +0.0069] & $-0.0006$ [-0.0104, +0.0091] \\
\bottomrule
\end{tabular}
\caption{Adjusted raw-only-diametric effects on excess transfer error. Entries are coefficient estimates with 95\% cluster-robust confidence intervals. The first analysis controls for released-label density and focal target-label presence; the second compares only observations with the same complete released five-label pattern.}
\label{tab:compression-controls}
\end{table*}

\begin{figure*}[t]
\centering
\includegraphics[width=0.75\textwidth]{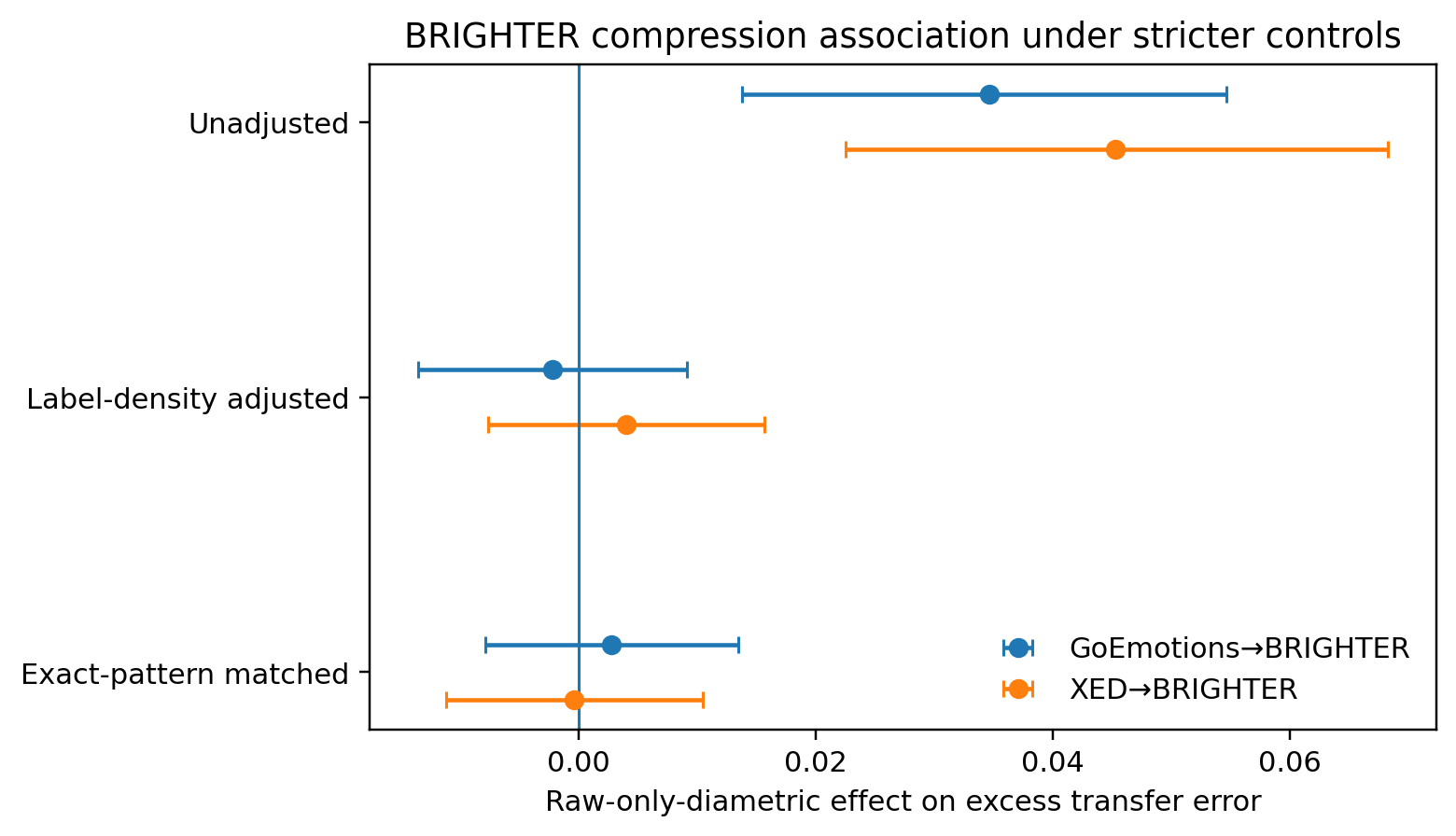}
\caption{Raw-only-diametric effect on uncalibrated excess transfer error under progressively stricter controls. Error bars are 95\% confidence intervals. The positive unadjusted association disappears after adjustment for released label structure and under exact five-label-pattern matching.}
\label{fig:compression-robustness}
\end{figure*}

The controlled analyses do not support an independent compression effect (Table~\ref{tab:compression-controls};
Figure~\ref{fig:compression-robustness}). For the raw predictions, the pooled coefficient after adjustment for released label density is $+0.0009$ [-0.0098, +0.0115], and the pooled exact-pattern coefficient is $+0.0012$ [-0.0087, +0.0111]. Effects after Platt and isotonic calibration are likewise close to zero. The initial association therefore appears to reflect the released label configurations in which raw-only diametric cases occur rather than information loss from annotation compression itself.

This analysis also does not account for the residual \XED{}--\BR{} \lab{surprise} divergence. For \lab{surprise}, raw-only items show no additional residual error after recalibration: the Platt differences are $+0.0063$ [-0.0193, +0.0312] for
\GE{}$\rightarrow$\BR{} and $-0.0104$ [-0.0367, +0.0156] for \XED{}$\rightarrow$\BR{}, with similarly null isotonic differences
($+0.0054$ [-0.0248, +0.0353] and $-0.0151$ [-0.0432, +0.0132], respectively). We therefore treat annotation compression as a tested but unsupported explanation for the ordering divergence reported in the main text.

\section{In-Domain Corpus Comparisons}
\label{app:indomain}

Table~\ref{tab:measurement-flips} gives the three corpus-pair--label comparisons from Section~\ref{sec:measurement} whose direction still depends on the head after rank normalization. Because the compared corpora are also the heads' training domains, these reversals show that measurements depend on the training resource, not how often reversals occur on external corpora.

\begin{table}[t]
\centering
\small
\setlength{\tabcolsep}{4pt}
\begin{tabular}{lrrr}
\toprule
Comparison & \BR{} & \GE{} & \XED{} \\
\midrule
BR--XED fear    & +.082 & +.069 & -.051 \\
BR--XED sadness & +.043 & +.109 & -.026 \\
GE--XED joy     & +.021 & +.018 & -.015 \\
\bottomrule
\end{tabular}
\caption{Descriptive rank-normalized corpus differences under each head, averaged across three seeds; item-sampling intervals are not estimated here. Positive values indicate that the first corpus has the higher mean pooled percentile rank; negative values indicate the reverse.}
\label{tab:measurement-flips}
\end{table}

\section{Complete External Collection Comparison}
\label{app:external}
Table~\ref{tab:external-full} reports the complete external comparison, with differences defined as 2016 minus 2017. We use 1,000 item-bootstrap resamples within each collection, sharing draws across heads and recomputing pooled ranks in each resample. Brackets in the interval columns give the resulting pointwise 95\% intervals, which quantify item-sampling uncertainty conditional on the fitted, seed-averaged predictions; seed ranges are minimum and maximum differences across seeds 13, 42, and 87. These are distinct sources of variability.
\begin{table*}[t]
\centering
\scriptsize
\setlength{\tabcolsep}{3pt}
\begin{tabular}{llrrrrrr}
\toprule
Label & Head & Raw & 95\% interval & Seed range & Rank & 95\% interval & Seed range \\
\midrule
\textit{anger} & \BR{} & +0.013 & [+0.001, +0.024] & [+0.010, +0.014] & +0.011 & [-0.000, +0.021] & [+0.004, +0.016] \\
\textit{anger} & \GE{} & +0.005 & [+0.001, +0.009] & [+0.005, +0.005] & +0.012 & [-0.000, +0.023] & [+0.007, +0.015] \\
\textit{anger} & \XED{} & +0.003 & [-0.000, +0.006] & [+0.003, +0.004] & +0.001 & [-0.010, +0.011] & [-0.003, +0.003] \\
\textit{fear} & \BR{} & +0.043 & [+0.030, +0.056] & [+0.037, +0.048] & +0.038 & [+0.026, +0.050] & [+0.034, +0.041] \\
\textit{fear} & \GE{} & +0.005 & [+0.000, +0.009] & [+0.003, +0.006] & +0.009 & [-0.003, +0.020] & [+0.005, +0.010] \\
\textit{fear} & \XED{} & +0.005 & [+0.001, +0.009] & [+0.002, +0.006] & +0.006 & [-0.005, +0.018] & [-0.005, +0.009] \\
\textit{joy} & \BR{} & -0.052 & [-0.067, -0.037] & [-0.057, -0.046] & -0.036 & [-0.048, -0.025] & [-0.038, -0.033] \\
\textit{joy} & \GE{} & -0.013 & [-0.019, -0.007] & [-0.018, -0.009] & -0.046 & [-0.058, -0.034] & [-0.048, -0.041] \\
\textit{joy} & \XED{} & -0.021 & [-0.027, -0.015] & [-0.028, -0.014] & -0.050 & [-0.062, -0.038] & [-0.053, -0.042] \\
\textit{sadness} & \BR{} & -0.001 & [-0.014, +0.010] & [-0.002, -0.000] & +0.005 & [-0.006, +0.016] & [+0.001, +0.007] \\
\textit{sadness} & \GE{} & -0.006 & [-0.013, +0.000] & [-0.007, -0.006] & -0.000 & [-0.011, +0.010] & [-0.002, +0.000] \\
\textit{sadness} & \XED{} & -0.004 & [-0.009, +0.000] & [-0.006, -0.003] & -0.008 & [-0.019, +0.003] & [-0.012, -0.005] \\
\textit{surprise} & \BR{} & +0.023 & [+0.013, +0.034] & [+0.019, +0.025] & +0.024 & [+0.012, +0.036] & [+0.020, +0.027] \\
\textit{surprise} & \GE{} & +0.001 & [-0.002, +0.004] & [+0.001, +0.001] & -0.003 & [-0.014, +0.009] & [-0.006, -0.002] \\
\textit{surprise} & \XED{} & +0.001 & [-0.001, +0.003] & [-0.000, +0.002] & -0.015 & [-0.026, -0.004] & [-0.024, -0.002] \\
\bottomrule
\end{tabular}
\caption{Complete external comparison under all three heads.}
\label{tab:external-full}
\end{table*}

\section{Exploratory LLM-Assisted Annotation Revision}
\label{app:llm-revision}

We tested whether revising suspicious training annotations improved native-specialist performance. An exploratory audit used  \texttt{gpt-5.6-sol} to assess 240 items, comprising 180 flagged candidates and 60 unflagged controls, with 80 items per resource. Candidates were selected using annotation and model-based audit signals. The sample was deliberately enriched for suspicious cases and is not representative of annotation-error prevalence.

The LLM received each item's text, assigned labels, and available label inventory, but not its dataset identity, audit scores, or model predictions. It identified unsupported assigned labels and strongly plausible missing labels. For this experiment, we revised only flagged training items, excluding judgments marked as having insufficient context. Unsupported labels were removed and suggested missing labels were added within the resource's observed
inventory. Development and test annotations remained unchanged. The revision changed 10 training items in \BR{},
31 in \GE{}, and 21 in \XED{}, involving 13, 64, and 47 binary label changes, respectively.

We retrained the three native specialists using the original training recipe and seeds 13, 42, and 87. Thresholds were selected on the unchanged development sets, and evaluation used the unchanged test sets. Table~\ref{tab:llm-correction} reports paired changes relative to the original specialists.

\begin{table}[t]
\centering
\small
\setlength{\tabcolsep}{4pt}
\begin{tabular}{lrrr}
\toprule
Resource
& $\Delta$ Macro-F1
& $\Delta$ Micro-F1
& $\Delta$ Brier \\
\midrule
\BR{}
& \meansd{-.00690}{.02156}
& \meansd{-.00772}{.01980}
& \meansd{+.000236}{.000549} \\
\GE{}
& \meansd{+.00230}{.00125}
& \meansd{+.00365}{.00547}
& \meansd{-.000025}{.000043} \\
\XED{}
& \meansd{-.00143}{.00550}
& \meansd{+.00135}{.00231}
& \meansd{+.000171}{.000177} \\
\bottomrule
\end{tabular}
\caption{Native test-set changes after LLM-assisted revision of flagged training annotations. Changes are revised minus original; parentheses give sample SD of paired changes across three seeds. Positive F1 and negative Brier changes indicate improvement.}
\label{tab:llm-correction}
\end{table}

Changes were small and mixed. \GE{} improved slightly on all three metrics, whereas \BR{} worsened and \XED{} showed mixed F1 changes with slightly worse Brier. This pilot therefore yielded no consistent benefit across resources. It does not establish that the original annotations were correct, rule out benefits from more extensive revision, or test an LLM used directly as an emotion predictor.
\end{document}